\documentclass[letterpaper, 10 pt, conference]{article}  
\usepackage{graphicx} 
\usepackage{color}
\usepackage{amsmath}
\usepackage{amssymb}  
\usepackage{underscore}
\usepackage{fancyvrb}
\usepackage{algorithmic}

\title{Model-Based Testing, \\Using Belief-Desire-Intentions Agents, \\of Control Code for Robots \\in Collaborative Human-Robot Interactions}

\author{Dejanira Araiza-Illan, Tony Pipe and Kerstin Eder\footnote{This work was supported by the EPSRC grants EP/K006320/1 and EP/K006223/1, part of the project ``Trustworthy Robotic Assistants''. 
Dejanira Araiza-Illan and Kerstin Eder are with the Depart\-ment of Computer Science and the Bristol Robotics Laboratory, University of Bristol, UK. E-mail: {\tt\footnotesize \{dejanira.araizaillan, kerstin.eder\}@bristol.ac.uk}. 
Tony Pipe is with the Faculty of Engineering Technology and the Bristol Robotics Laboratory, University of the 
West of England, Bristol, UK. E-mail:~{\tt\footnotesize tony.pipe@brl.ac.uk}}
}
\date{}

\begin{document}

\maketitle
\thispagestyle{empty}
\pagestyle{empty}

\begin{abstract}
The software of robotic assistants needs to be verified, to ensure its safety and functional correctness. 
Testing in simulation allows a high degree of realism in the verification. 
However, generating tests that cover both interesting foreseen and unforeseen scenarios in human-robot interaction (HRI) tasks, while executing most of the code, remains a challenge. 
We propose the use of belief-desire-intention (BDI) agents in the test environment, to increase the level of realism and human-like stimulation of simulated robots. 
Artificial intelligence, such as agent theory, can be exploited for more intelligent test generation. 
An automated testbench was implemented for a simulation in Robot Operating System (ROS) and Gazebo, of a cooperative table assembly task between a humanoid robot and a person. 
Requirements were verified for this task, and some unexpected design issues were discovered, leading to possible code improvements.  
Our results highlight the practicality of BDI agents to automatically generate valid and human-like tests to get high code coverage, compared to hand-written directed tests, pseudorandom generation, and other variants of model-based test generation.
Also, BDI agents allow the coverage of combined behaviours of the HRI system with more ease than writing temporal logic properties for model checking. 

\end{abstract}

\section{INTRODUCTION}

Interactions between humans and robots at home, at work, in hospitals and on the street, have ethical and legal implications. 
As robot designers at software level, we must ensure the safety of people, the environment and the robotic systems, beyond just restricting the operation of robots to stop them when humans get close~\cite{ROMAN14}. 
Furthermore, demonstrating safety and functional soundness of robotic technologies opens the doors for them to become viable commercial products.

Robotic code needs to be verified to eliminate runtime bugs, such as floating point issues and memory allocation, as in~\cite{Trojanek2014}. 
Additionally, robotic code must be verified and validated when it interacts with other related code, the robot's hardware components, the environment and people. All these elements introduce concurrency, and thus the emergence of possible unexpected and undesirable behaviours. 
A challenge to face in the verification and validation of any robotic platform is the interaction with complex, multifaceted, dynamic and evolving environments~\cite{Micskei2012,CDV2015}. 
Consequently, verifying robotic software goes beyond introducing correct data to stimulate the code, the classical software verification research focus~\cite{Nie2011}.
Meaningful and realistic orchestration of sequences of data inputs that simulate interesting scenarios is necessary for testing robots in human-robot interactions (HRI)~\cite{Mossige2014}. How can we achieve realism in the orchestration, considering such complex environments?

Formal methods provide proofs for safety and functional soundness requirement satisfaction or violation.  
Most of these approaches require abstract models of the robot's elements (software and hardware) and its target environment (including people). 
A challenge is the formulation of adequate models, that capture sufficient detail to be meaningful, whilst being computationally manageable at the same time. 
Current model checking approaches for robotic software verification require a great deal of abstractions from the physical world, often focusing on a decision-making level only (e.g.,~\cite{webster14formalshort,Dennis2015}). 
Model checking tools for code target specific programming languages such as ANSI-C\footnote{http://www.cprover.org/cbmc/}, C++\footnote{http://divine.fi.muni.cz/} and Java\footnote{http://javapathfinder.sourceforge.net/}. 
Nonetheless, other robotic system components such as sensors and actuators, and environments with people in HRI, ought to be considered in combination with the code.

Testing can be performed on models of robotic systems and their environments in simulations~\cite{Pinho2014}, for real robots in the real world (as in~\cite{Mossige2014}), or on combinations of both~\cite{Petters2008}. 
In simulation, models can include a great level of detail about the physical world in which HRIs take place. Real control code can be coupled with simulated physics, sensing and actuating~\cite{Pinho2014,CDV2015}. 
The generation of effective tests is the main challenge of testing, i.e., targeting meaningful and interesting scenarios, while at the same time, exploring the state space to discover unexpected bugs (e.g., covering all the code, most of the possible values of a variable, or most types of scenarios). 
Random sampling methods, such as rapidly-exploring random trees~\cite{Kim2006}, allow wide exploration.
Constraints focus the test generation to create meaningful and valid tests~\cite{Mossige2014}. 
Alternatively, model-based test generation allows the substitution of constraint solving by model analysis methods (e.g., model checking).

Although many languages and formalisms have been proposed for model-based test generation (e.g., UML and process algebras for concurrency~\cite{Lill2012}, or Lustre and MATLAB/Simulink for data flow~\cite{Utting2012}), none of these are suitable to represent complex environments with people, such as HRI scenarios. Is it possible to use agent-based models, such as belief-desire-intention (BDI), for test generation, to achieve more realistic human-like behaviour when stimulating a robot? 
BDI agents are already used to model tasks of planning and decision-process mechanisms, with the added benefit of analysis via model checking (e.g.~\cite{Bordini2006}).

In microelectronics, Coverage Driven Verification (CDV) is a method that guides the generation of effective tests, according to feedback from coverage metrics, and automation in the testing process~\cite{Pizialli2004}. 
In~\cite{CDV2015}, we illustrated the implementation of automatic CDV testbench components (test generation, driver, checker, coverage collection) to verify robotic code for HRI within a simulator developed within the robotics programming framework of Robotic Operating System (ROS). 
We experimented with unconstrained-pseudorandom, constrained-pseudorandom and model-based test generation, to assess their strengths empirically. 
The model-based approach consisted of two steps, where abstract test sequences are generated first via model checking of probabilistic timed automata (PTA) in UPPAAL\footnote{http://www.uppaal.org/}. Subsequently a refinement step transforms them into concrete test sequences, according to realistic models. These concrete sequences are employed to stimulate the environment, sensors and actuators in the robot, which will generate inputs for the robotic code under test, i.e.\ we stimulate the robot's code through exposure from its environment.

The main contribution of this paper is the exploitation of belief-desire-intention (BDI) agents in testing environments, for model-based test generation. Through BDI test generation, we gain more realistic human-like stimulus to verify robotics code in simulation, and simplify the task of generating interesting test scenarios. 
Additionally, we illustrate BDI-based test generation on a sophisticated HRI case study.

We applied testing in simulation to a scenario consisting of the cooperative human-robot manufacture of a table, where the robot hands out the legs, and the human attaches them to the table top. 
A simulator of the table manufacture scenario was developed in ROS-Gazebo, coupled to an automated testbench in ROS. 
We included constrained-pseudorandom and model-based test generation, in a two-tiered approach, as we have found they complement each other to explore the code under test and particular scenarios of interest as per our requirements to verify~\cite{CDV2015}. 
We tested the high-level table assembly robotic control code, to verify safety and functional requirements. 
Our results illustrate that the code under test, and many combinations of the environment and the robotic system, can be covered in exploration through the use BDI agents. 
The use of BDI agents is a viable alternative for model-based test generation.

Our results highlight the potential of \textit{(a)} using BDI agents for model-based test generation to stimulate robotics code according to realistic scenarios; and \textit{(b)} having a framework for testing your robotic code in HRI simulators, for a cost-effective software development process.

The paper proceeds as follows. 
Section \ref{sc:casestudy} introduces the table assembly task scenario. 
Section \ref{sc:testbench} presents an overview of the CDV testbench components, with emphasis on our proposed BDI-based test generation approach. 
Section \ref{sc:experiments} discusses the coverage and fault discovery results. 
Section \ref{sc:relatedwork} reviews related work on test generation for robotic systems. 
We conclude the paper in Section \ref{sc:conclusion}.

\section{COOPERATIVE MANUFACTURE TASK}\label{sc:casestudy}

\begin{figure}[t]
\centering
\includegraphics[width=0.475\textwidth]{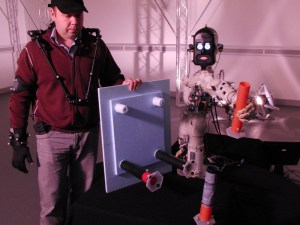}
\includegraphics[width=0.41\textwidth, trim=17cm 8cm 15cm 5cm,clip]{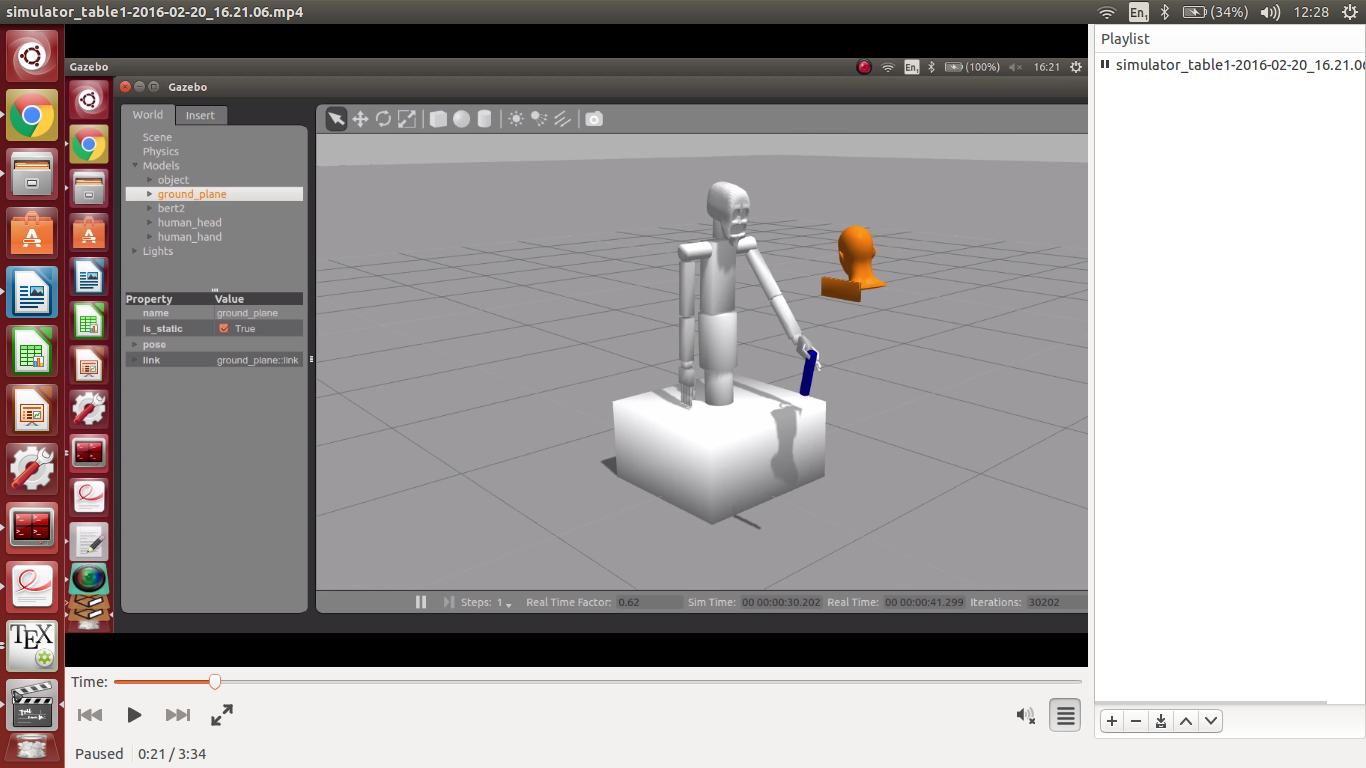}
\caption{BERT2 robot, table to assemble, and their simulation versions}
\label{table}
\end{figure}

Our case study is the cooperative assembly of a table consisting of 5 components, 4 legs and the table top, as shown in Fig.~\ref{table}. 
The robot, BERT2~\cite{lenz2010bert2} (also in Fig.~\ref{table}), should hand over legs to the person, one at a time, when the person asks the robot and only if the robot decides the human is ready. 
The robot keeps count of the number of legs the person has taken, reporting that a table has been completed if four legs have been handed over within a time threshold. 

The handover of the legs is the critical part in this scenario.
In a successful assembly, a handover starts with a voice command from the person to the robot, requesting a table leg.  
The robot then picks up a leg, holds it out to the human, and signals for the human to take it. 
The human issues another voice command, indicating readiness to take the leg. Then, the robot makes a decision to release the leg or not, within a time threshold, based on a combination of three sensors: ``pressure,'' indicating whether the human is holding the leg; ``location,'' visually tracking whether the person's hand is close to the leg; and ``gaze,'' visually tracking whether the person's head is directed towards the leg. 

The sensing combination is the Cartesian product of ``gaze'', ``pressure'' and ``location'' readings, $(G,P,L)\in G \times P \times L$. Each sensor reading is classified into $G,P,L=\{\bar{1},1\}$, $1$ indicating the human is ready according to that sensor, and $\bar{1}$ for any other value. 
If the human is deemed ready, $GPL=(1,1,1)$, the robot should decide to release the leg. Otherwise, the robot should not release the leg and discard it (send back to a re-supply cycle). 
The robot will time out while waiting for either a voice command from the human, or the sensor readings, according to specified time thresholds. 

A ROS `node' in Python, with 264 statements implements the high-level robot control of the assembly task, reusing as much as the handover code in~\cite{CDV2015} as possible. The code was structured as a finite-state machine (FSM) using the SMACH modules~\cite{SMACH}, as it allows to extract abstract models of the code in a semi-automatic manner. 

\subsection{Requirements List}\label{ssc:requirements}

We considered a selected set of safety and functional requirements, for illustrative purposes, adapted from the handover example in~\cite{CDV2015}:

\begin{enumerate}
\item If the gaze, pressure and location sense the human is ready, then a leg shall be released (functional). 
\item If the gaze, pressure or location sense the human is not ready, then a leg shall not be released (functional). 
\item The robot shall not close its hand when the human is too close (safety).
\item The robot shall start and work in restricted joint speed of less than 0.25 rad/s (safety). 
\end{enumerate}

\subsection{ROS-Gazebo Simulator and Models}
The ROS-Gazebo simulator, available online\footnote{https://github.com/robosafe/table}, includes: 
\textit{(a)} the robotic code under test corresponding to the high-level control that enacts the table assembly and object handover protocol; 
\textit{(b)} physical models of the robot, human and objects, with the physics of our world (gravity, collisions) in Gazebo, where the elements move by updating their pose; 
\textit{(c)} sensor models for pressure, gaze, location and voice commands, with stochastically modelled errors due to the object wobbling in the robot's hand (pressure error), occlusions (location and gaze errors), and noise (voice command errors); 
and \textit{(d)} other code such as the kinematic trajectory planner in MoveIt!\footnote{http://moveit.ros.org/}.
We improved upon the requirements verification results in~\cite{CDV2015}, implementing speed limits for safety to comply with the 0.25 rad/s limit.

\section{TESTBENCH COMPONENTS}\label{sc:testbench}

We implemented testbench components: test generator, driver, checker, and coverage collection, for the table assembly case study, based on our previous work~\cite{CDV2015}. 
The modularity of the testbench components and the simulator facilitates re-usability, when elements are improved or modified to reflect design changes, and extendibility, e.g., increasing the degree of detail or abstraction.

\subsection{Test Generation}

\begin{figure}[t]
\centering
\includegraphics[width=0.8\textwidth]{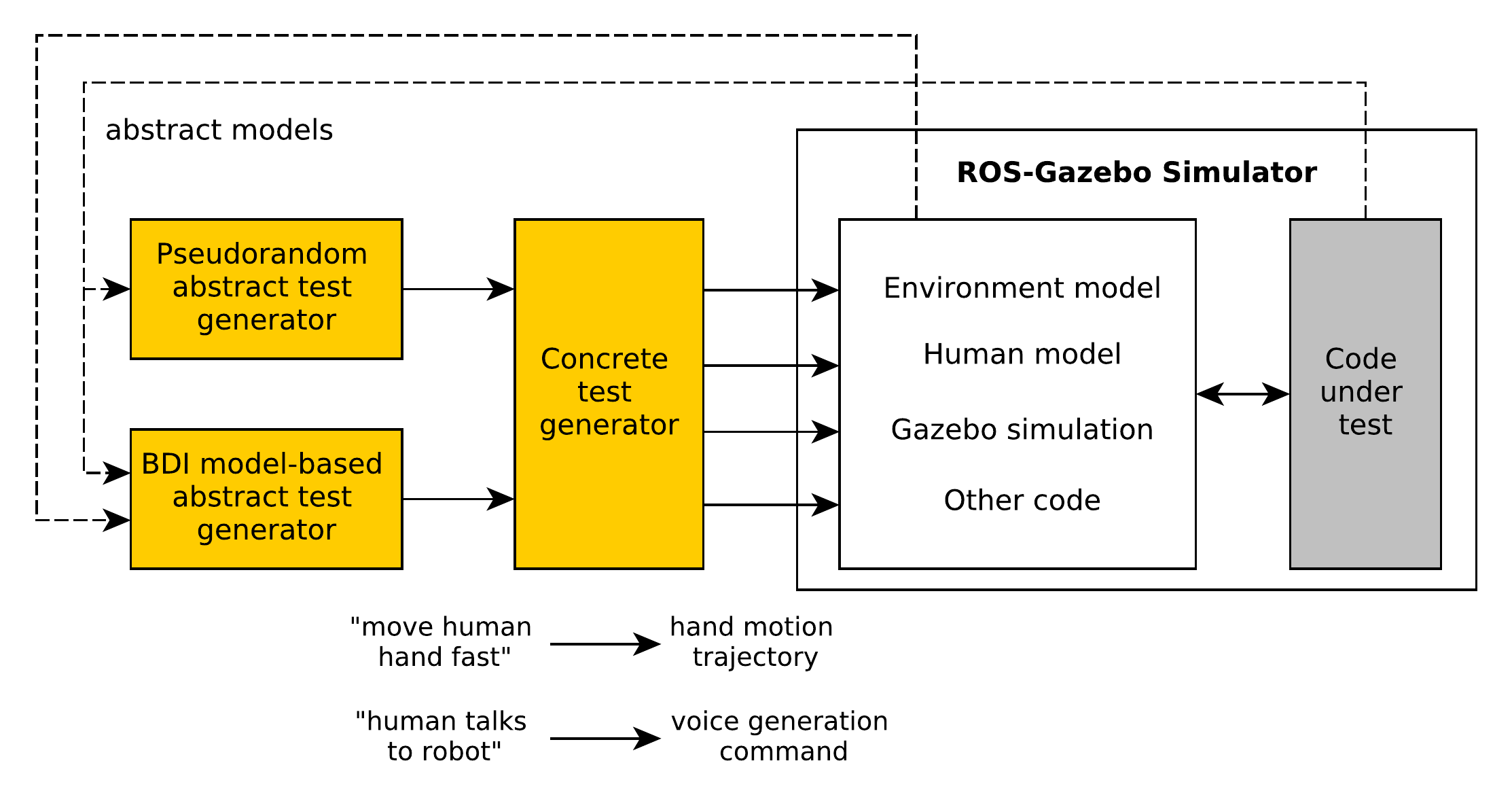}
\caption{Two tiered model-based test generation approach}
\label{Testgenerationflow}
\end{figure}

We implemented constrained-pseudorandom and BDI-based test generation, through a two-tiered approach, illustrated in Fig.~\ref{Testgenerationflow}. 
This approach targets the stimulation of robotics code through models of its environment and other related software and hardware (sensors and actuators) in simulation, in a realistic manner. 
In contrast, most generic approaches in software choose the inputs to the code under test and apply them directly, which does not guarantee that the code's stimulation will be close to real-life stimuli, and can easily create situations that would not occur in practice. 
We synthesize an abstract test sequence for the environment, sensors and actuators, which will stimulate the robotic code in the interaction. 
These abstract actions come from models of the robotic code under testing, in combination with the environment's possibilities.
The abstract sequence is used by the simulator to assemble code fragments, so that the human and the environment enact the test accordingly.
Subsequently, a concrete generator is employed for the instantiation of parameters used by the assembled code, from legal and predefined sets of the state space (e.g., equivalence class partitioning~\cite{Gao2014}), via constraint solving, search methods or random sampling~\cite{Gaudel2011}. 
This process allows stimulating the human and environment components in the simulator to timely emulate motion in the real world. 
Similar tiered abstraction-refinement processes can be found in control systems synthesis applications, such as~\cite{Fainekos2005,Nenchev2015}. 
For example, a ``move human hand fast'' sequence element is sent to the hand motion concrete generator, which computes a motion trajectory for the hand in the 3D space under the ``fast'' constraint, from the valid subset of the state space; then, ``human talks to robot'' is sent to the corresponding concrete generator to select an input for the voice sensor.

Abstract models can be derived semi-automatically from the robot's code, from structuring it into finite-state machines (e.g., using the SMACH Python module~\cite{SMACH}). 
Furthermore, the concrete parameter generators can be replaced, without modifying the abstract models. 
Thus, our approach facilitates reuse of components and adaptation, due to its modularity.

In unconstrained, pseudorandom test generation, all the test inputs for stimulating the code are chosen from sampling their domains, from carefully chosen probability distributions to maximize coverage and fault detection~\cite{Gaudel2011}. 
Constraint solving and search methods have been employed to bias testing towards valid and rare events, and specific kinds of scenarios for the robotic systems~\cite{Kim2006,Mossige2014}. Model-based approaches ``argue'' that biasing can be performed in a more systematic and automatic manner. 
For our two tiered-approach, valid abstract test sequences are concatenated from the possible alphabet of environmental and human actions in the HRI task, with uniform probability. 
Constraints are used to fix a particular sequence element, or to indicate order relationships between the elements. 
This allows a deviation from a coherent and rational human and environmental behaviour strictly following the HRI protocol, to check that the code under test is robust. 

In model-based test generation, model checking (e.g. in~\cite{CDV2015}) or exploration of a model is performed, to synthesize a test sequence that complies with a property or specification~\cite{Utting2012}.  
In this paper, we use BDI agents to model the desired functionality of the human and the robot in the HRI task, a table assembly.

BDI is an intelligent or rational agent architecture for multi-agent systems (MAS). BDI models human practical reasoning, proposed by the philosopher Michael Bratman, which focuses on `intentions' or a list of future plans to achieve a goal~\cite{Agentspeakbook}. 
Different software frameworks can be employed for BDI agent development, such as Jason\footnote{http://jason.sourceforge.net/wp/}, many of these implemented in Java. 
Jason interprets agents expressed in the AgentSpeak language. 
An agent is defined by initial `beliefs' (the initial state of the agent) as (first-order) atomic formulae, and a set of `plans' forming a plan library. The agent has `goals' (target states of the world). 
A plan has a `head', i.e., a triggering event and an expression about beliefs (a `context'), and a `body' or set of actions to execute. 
Plans are triggered and executed following `events', e.g., new beliefs or new goals, according to the head. 
New beliefs are caused by other agents, or by the execution of plans (self-beliefs)~\cite{Bordini2005}.

BDI agents are an ideal candidate to model rational human-like decision making, based on beliefs and goals, in HRI. 
Consequently, we propose to model the human using BDI, along with the environment and robotic system, for test generation.

In our two-tiered test generation approach (abstract to concrete), the employed BDI agents are abstractions of the actions and the robot's code FSM structure. 
To generate abstract test sequences for the robot's code, plan execution can be biased through controlling the beliefs, the goals, or both simultaneously, of the agents in the environment, e.g., modifying what the human is perceiving, which will trigger a plan that changes the world the robot is exposed to, thus triggering robot actions.
The control of the agents can be achieved by a `meta' agent supplying beliefs (or beliefs and goals) according to coverage metrics to explore all the environmental actions, and accumulating metrics about the robot's code coverage (from the robot's code agent).

The pseudo-algorithm for guiding the `meta' agent to control the environment agents is provided in Fig.~\ref{fig:pseudocode}. 
The architecture of the human and environment agents is known beforehand, and so are the beliefs and goals that trigger different plans. 
Data structures of beliefs and goals to manipulate are explored, selecting which ones will be triggered in the human or environment in a run of the BDI multi-agent system. 
Once a run is finished, coverage statistics can be gathered about the exploration of the robot's code and the environment agents, to modify the next exploration of beliefs and goals. 

Each run of the multi-agent system will produce an abstract test sequence. 
The production of abstract test sequences will stop when the robot's code coverage is within an acceptable threshold, or when the environment agents have been fully explored. 
Note that we assume that the environment agents do not have infinite loops. 

\begin{figure}[t]
\centering
\footnotesize
\begin{algorithmic}[1]
\STATE $\mathcal{E}=$ list of beliefs and goals of the environment agents (e.g., \texttt{human})
\WHILE{cov(\texttt{robotcode})$\neq$total \OR cov(environmentAgents)$\neq$total} \STATE{Choose beliefs (or goals, or both) from $\mathcal{E}$, for \texttt{meta} agent} \STATE {Run multi-agent system} \STATE{Write abstract test sequence into ROS} \STATE{Collect coverage of environment agents} \STATE{Collect coverage of robot's code}
 \ENDWHILE
\end{algorithmic}
\caption{Algorithm to produce abstract test sequences via BDI agents}
\label{fig:pseudocode}
\end{figure}

An example of plans of \texttt{meta}, \texttt{human} and \texttt{robotcode} agents is shown in Fig.~\ref{fig:agents}. In this example, a belief \texttt{leg2} is added in the agent \texttt{human} by the agent \texttt{meta}, which triggers the plan starting with \texttt{+!activate:leg2}. This plan triggers a belief \texttt{leg} in the agent \texttt{robotcode}, which activates the plan \texttt{+!waiting:leg}. 
 
\begin{figure}[t]
\centering
\footnotesize
\begin{Verbatim}[frame=single,numbers=left,numbersep=2pt,xleftmargin=0.25cm]
//Agent meta
/* Initial beliefs and rules */
/* Initial goals */
/* Plans */
+!control : true <- .send(human,tell,leg2).
...
\end{Verbatim}

\begin{Verbatim}[frame=single,numbers=left,numbersep=2pt,xleftmargin=0.25cm]
//Agent human
/* Initial beliefs and rules */
/* Initial goals */
!activate.
/* Plans */
+!activate : leg2 <- .send(robot_code,tell,leg).
...
\end{Verbatim}

\begin{Verbatim}[frame=single,,numbers=left,numbersep=2pt,xleftmargin=0.25cm]
//Agent robotcode
/* Initial beliefs and rules */
/* Initial goals */
!waiting.
/* Plans */
+!waiting : not leg <- !waiting.
+!waiting : leg <- !grabLeg. 
...
\end{Verbatim}

\caption{Human and robot's code BDI agent snipets in AgentSpeak}
\label{fig:agents}
\end{figure}

For the case study, we manually selected which beliefs in the \texttt{human} agent would be used for its manipulation, from all the available ones in the agent description, e.g., number of legs to request, becoming bored, or setting up gaze, pressure and location. 
Selected combinations of these beliefs were triggered automatically in the BDI model to generate abstract test sequences, to systematically explore the code under test, as well as combinations of interesting behaviours of the human and the robot in the interaction.

Model-based test generation through model checking PTA in UPPAAL~\cite{CDV2015} entails the need for an abstract model of the system and the HRI protocol requirements, such that it is traversable; i.e., a model that avoids the state-space explosion problem due to a large number of variables with several possible values. 
These time consuming model abstraction processes are often iterated by hand. 
Furthermore, reachability properties ought to be formulated for the model checker to produce an abstract test, which requires a good understanding of formal logics. 

\subsection{Driver}

The driver routes each one of the concrete inputs to the respective simulator component (e.g., the motion trajectory is fed to Gazebo to enact it on its 3D world representation), to stimulate the robot through these environment, sensor and actuator components. 

\subsection{Checker}

We followed our approach of implementing assertion monitors as automata to check each one of the requirements, using the SMACH module in Python~\cite{CDV2015}. 
The assertion monitors refer to different types of ``events'': \textit{(a)} abstract (e.g., the robot moved and the human responded); \textit{(b)} ``continuous'' parameters such as the robot's speed; \textit{(c)} code-related (e.g., function \texttt{foo} was triggered; and \textit{(d)} combinations of any of the previous. 

Automata monitors are spawned every time the triggering event occurs, and the subsequent checks ensue. 
The results of the checks are recorded for coverage collection. 
The automata can be spawned once until a change of events (i.e., a new spawning occurs when the triggering event is true again, after some time of it being false), or they can be spawned every specified $\Delta t$ time as long as the triggering signal is true.

\subsection{Coverage Collection}\label{ssc:coverage}

We implemented code statements coverage over the Python node under test, instrumented via the `coverage' Python module\footnote{http://coverage.readthedocs.org/en/coverage-4.1b2/}~\cite{CDV2015}. 
We also implemented assertion coverage, assessing if an assertion monitor is triggered per test run~\cite{CDV2015}. Ideally, all the assertions must be covered by our test suite.
Finally, we implemented cross-product or Cartesian product coverage over critical abstract action tuples $\langle Human,Robot\rangle$, such as four successful leg requests that were handed over, occasions when the human got bored and the robot discarded the leg, or occasions when the human acted but never sent the first voice command asking for a leg. 
Ideally, each tuple is encountered at least once within a test suite.

\section{EXPERIMENTS AND RESULTS}\label{sc:experiments}

We tested the robots code in the table assembly task, to verify the requirements in~\ref{ssc:requirements}. 
The simulator and testbench were implemented in ROS Indigo and Gazebo 2.2.5. The tests were run on a PC with Intel i5-3230M 2.60\,GHz CPU, 8\,GB of RAM, running Ubuntu 14.04. We used Jason 1.4.2 for BDI-based test generation.

Coverage was collected according to the models in Section~\ref{ssc:coverage}, when using model-based and constrained-pseudorandom test generation methods. 
We generated 130 abstract tests from a possible total number of 32768, by constraints over the possible beliefs for the \texttt{human} agent in the BDI multi-agent model, provided by the \texttt{meta} agent. 
In particular, we sought to cover combinations of human actions to set up the gaze, pressure and location sensor readings, such that \textit{(a)} $GPL=(1,1,1)$ for all the leg handover sub-tasks in a test; \textit{(b)} $GPL$ is not  $(1,1,1)$ in at least one of the leg handover sub-tasks in a test; \textit{(c)} the human successfully requested 4, 3, 2 or 1 legs in total per test (i.e., the protocol was followed without the human being bored at least 4, 3, 2 or 1 times); and \textit{(d)} legs were requested, but the human did not send the second readiness voice command or got bored. 
We generated a concrete test from each abstract sequence for \textit{(a)}-\textit{(c)} (128 concrete tests), and 5 concrete tests for each of the two abstract tests for \textit{(d)} (adding up to 138 tests in total), with seeds equal to the test numbers. 
Subsequently, we generated 30 abstract constrained-pseudorandom test sequences where \textit{(e)} no legs were requested at all, with one concrete test each and seeds equal to the test numbers. 
The abstract tests in \textit{(d)}-\textit{(e)} would cause the robot to time out. 
The robot is allowed to try a successful handover for a maximum of four times per test. 
Each test ran for a maximum of 300 seconds.

\subsection{Code Coverage}

\begin{figure}[!t]
\centering
\includegraphics[width=0.8\textwidth]{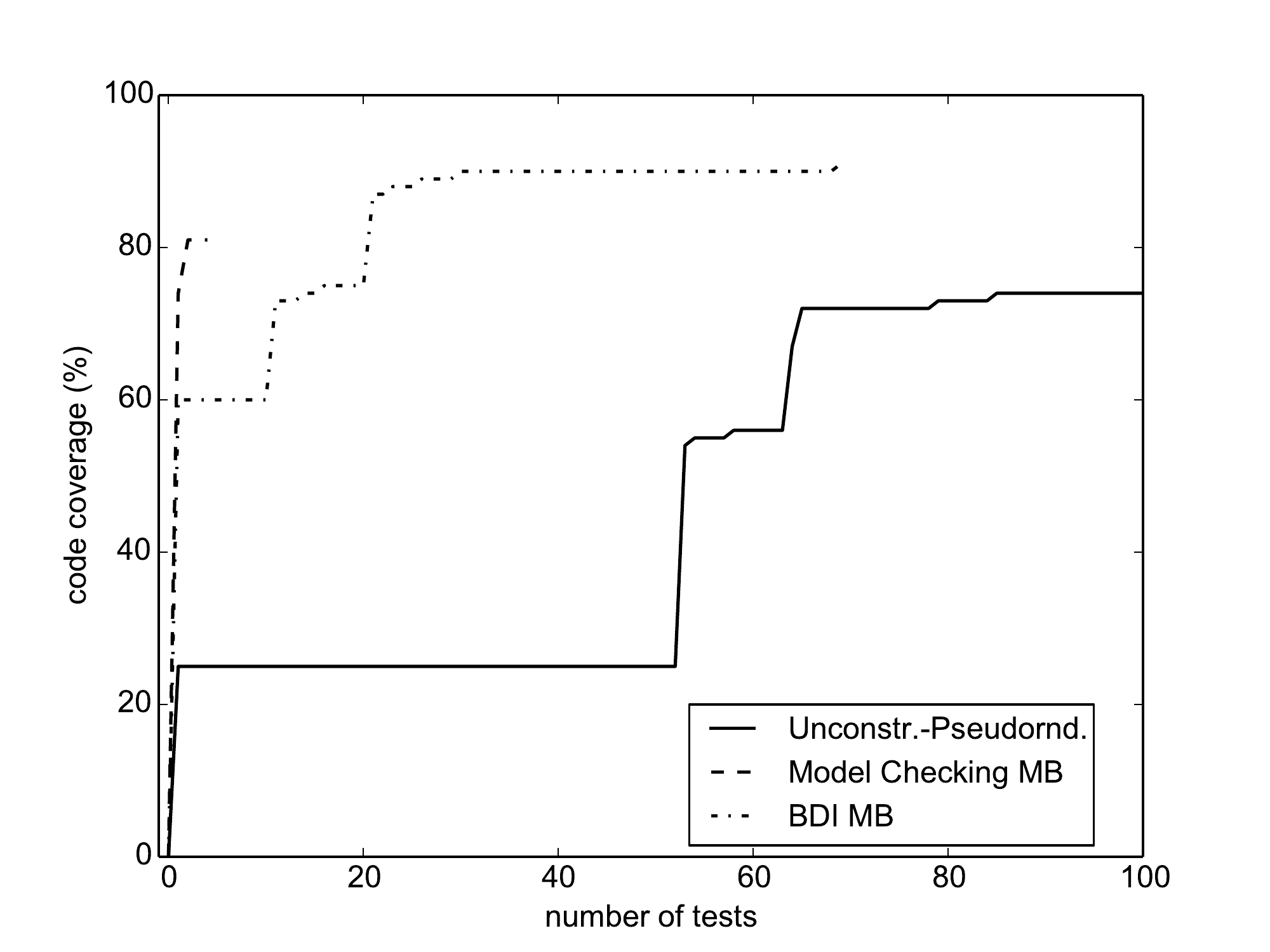}
\caption{Code coverage results}
\label{fig:code}
\end{figure}

The results in Fig.~\ref{fig:code} show that model-based methods are clearly better than randomized exploration to reach high levels of code coverage quickly. 
These maximum coverage percentages are reached when the robot decides to release or not to release an object to the person.

We reused the code coverage results from~\cite{CDV2015}, regarding model-based (model checking PTA in UPPAAL), and unconstrained, pseudorandom test generation. 
We selected some of the BDI-based test generation results, for a fair comparison of the code coverage regarding the handover part of the task.

The tests generated through model checking targeted reaching high levels of code coverage with a single test, by formulating this coverage requirement as temporal logic properties (for object release and no release).
With BDI agents, we also reached high levels of code coverage, by explicitly asking the \texttt{meta} agent to control the \texttt{human} agent to follow a handover to completion by setting 4 beliefs; i.e., the human would ask for a leg, then wait until the robot is ready, then set the head and hand, and finally indicate readiness to take the object. 
This process of setting beliefs to trigger plans was more intuitive than formulating temporal logic properties.

\subsection{Assertion Coverage}

The assertion coverage results are shown in Table~\ref{results1}. 
We recorded the number of tests where the requirement was satisfied (Passed), not satisfied (Failed) or not checked (NC). 

\begin{table}[!t]
\centering
\renewcommand{\arraystretch}{1.3}
\caption{Requirements (assertion) coverage}
\begin{tabular}{|c|c|c|c|c|c|c|}
\hline 
Req. & \multicolumn{3}{|c|}{Constrained-Pseudorandom} &\multicolumn{3}{|c|}{BDI Model-Based}  \\
	&  		Passed & Failed & NC		 & Passed & Failed & NC 			  \\
\hline
1 		& 0/30  & 0/30 & 30/30  	&  90/138 & 1/138 & 47/138 \\
2 		& 0/30  & 0/30 & 30/30 	&  100/138 & 0/138 & 38/138 \\
3 		& 0/30  & 0/30 & 30/30 	&  138/138 & 12/138 & 0/138 \\
4 		& 30/30 & 0/30 & --- 	&  138/138 & 0/138 & --- \\ 
\hline
\end{tabular} \label{results1}
\end{table}

Req.~2 was satisfied in all the tests, and Req.~1 was not satisfied only in one instance, as the release of the object took a longer time than the specified time threshold. 
Req. 3 is not satisfied in some tests, as the person's hand is allowed to get close to the robot gripper when it closes. 
To improve this issue, the person's hand would need to be tracked, and the robot gripper stopped accordingly.

Req. 4, inspired by the standard ISO~10218-1 for industrial robots, is satisfied in all tests.
From the observed results in~\cite{CDV2015}, speed thresholds were enforced in the kinematic planning executed by the MoveIt! package, which were obeyed by the system.

\subsection{Cross-Product Functional Coverage} 

Table~\ref{fig:cross-product} shows the coverage results for reachable tuples of interest, including when all requested legs were handed over, at least one leg was not handed over, the human was bored in a test and the robot discarded the leg, and the human never requested a  leg successfully. 
All the tuples were reached at least once in total. 
The results reflect the rareness of $GPL=(1,1,1)$ (tuples 1, 4 and 7) for all the requested legs in a run, considering errors in the sensor readings, and the likelihood of this $GPL$ combination compared to the other 7 possibilities.

\begin{table*}[!t]
\centering
\renewcommand{\arraystretch}{1.3}
\caption{Reachable cross-product coverage}
\begin{tabular}{|c|c|c|c|c|}
\hline 
\multicolumn{2}{|c|}{Tuple: $\langle Human,Robot \rangle$ }		& Constr.,P.rand.& BDI	& TOTAL \\
\hline
1&$\langle$4 legs, $GPL=(1,1,1)\times 4 \rangle$ 				& 0/30 			&  1/138		& 1/168 \\
2&$\langle$4 legs, $GPL\neq (1,1,1)$ for at least 1 leg$\rangle$ 	& 0/30  			&  15/138 	& 15/168 \\
3&$\langle$4 legs$+$bored, Sensing timed out$\rangle$ 				& 0/30 			&  16/138 	& 16/168 \\ \hline
4&$\langle$3 legs, $GPL=(1,1,1)\times 3$ legs $\rangle$ 			& 0/30 			&  1/138 	& 1/168 \\
5&$\langle$3 legs, $GPL \neq (1,1,1)$ for at least 1 leg$\rangle$	& 0/30 			&  15/138 	& 15/168  \\
6&$\langle$3 legs$+$bored, Sensing timed out$\rangle$ 				& 0/30 			&  16/138 	& 16/168 \\ \hline
7&$\langle$2 legs, $GPL=(1,1,1)\times 2$ legs $\rangle$ 			& 0/30 			&  2/138 	& 2/168 \\ 
8&$\langle$2 legs, $GPL \neq (1,1,1)$ for at least 1 leg$\rangle$	& 0/30 			&  15/138 	& 15/168 \\
9&$\langle$2 legs$+$bored, Sensing timed out$\rangle$ 				& 0/30 			&  14/138 	& 14/168 \\ \hline
10&$\langle$1 leg, $GPL=(1,1,1)\rangle$ 							& 0/30 			&  11/138 	& 11/168 \\ 
11&$\langle$1 leg, $GPL \neq (1,1,1)\rangle$ 					& 0/30 			&  22/138 	& 22/168 \\ 
12&$\langle$1 leg$+$bored, Sensing timed out$\rangle$ 				& 0/30 			&  5/138 	& 5/168 \\ \hline
13&$\langle$No leg, Timed out$\rangle$ 							& 30/30 			&  0/138 	& 30/168 \\ \hline
\end{tabular} \label{fig:cross-product}
\end{table*}

\subsection{Discussion}

The formulation of BDI agents allowed for models of the HRI task that represent rational human-like actions. 
Through this modelling scheme, the HRI protocol can be traversed exhaustively by controlling some of the agents via their beliefs. 
Compared to model checking, it is not necessary to formulate temporal logic reachability properties, which requires a good understanding of formal logics, and a greater degree of manual input. 
Furthermore, BDI agents are easier to program, by specifying plans of actions, instead of constructing PTA in UPPAAL. 
The latter needs several cycles of abstraction to deal with the state-space explosion problem, i.e. to be traversable by model checking.

Our approach moves model-based test generation towards automated coverage-directed generation~\cite{Pizialli2004}. 
Beliefs in the agents can be triggered to satisfy coverage goals, automated further by adding feedback loops in the algorithm of Fig.~\ref{fig:pseudocode}. 
The results in this section prove that high percentages of coverage were achieved quickly and efficiently using BDI agents, in less than 200 tests. 
Furthermore, ``directed tests'' can be assembled from the BDI agents, by introducing constraints for their beliefs.

\section{RELATED WORK} \label{sc:relatedwork}

Testing of robotic systems can be performed in a real-life setting~\cite{Mossige2014}, completely in simulation~\cite{Pinho2014}, or in combinations of simulation and real components~\cite{Petters2008}, i.e., hardware-in-the-loop. Our BDI approach offers a novel solution for the two latter cases. 
In our approach, we explore the code mainly for finding and eliminating functional bugs, i.e., for safety and functional soundness, although runtime bugs can be found by instrumenting the code with relevant assertion monitors. 

Test generation research has focused on applications where the tests have relatively small sets of data types, e.g., a timing sequence for controllers~\cite{Mossige2014}, producing images to verify image processing software\footnote{http://development.objectvideo.com/index.html}, or a set of state space inputs for a controller~\cite{Kim2006}. In our approach, the inputs to the simulator become combinations of these and several different types. Thus, our generation problem is much more complex and that is why we used the two tiered approach, from abstract to concrete tests.

Constraint solving requires mathematical models of the inputs to the system (code) to stimulate, in the form of constraint programs or optimization programs to solve~\cite{Mossige2014}. Some heuristics are needed to help the solvers, e.g., orders of variables.  
Search methods are an alternative to solve constraint programs or optimization problems~\cite{Mcminn2004,Kim2006,Dang2009,Sankaranarayanan2012}. Nevertheless, heuristics to guide the search are needed, e.g., cost functions. 
Hybrid systems approaches require the formulation of the test generation problem into a hybrid model (e.g., hybrid automata)~\cite{Julius2007,Dang2009}, which means a great deal of abstraction and manual input in practice. 

Other model-based approaches seek to test models at the same level of abstraction as the model-based test generation~\cite{Hessel2008}, or they focus on testing high-level functionality~\cite{Lill2012,webster14formalshort,Dennis2015}. Thus, test generation is much easier to implement than for our testing problem, which targets the real robotics code in HRI realistic scenarios. 
Our model-based test generation approach is based on divide-and-conquer, simplifying the constraint solving or search problem. This abstract-concrete process has been proposed for the synthesis of hybrid controllers that satisfy properties~\cite{Fainekos2005,Nenchev2015}.

To the best of our knowledge, BDI agents have not been used as the modelling formalism for model-based test generation before. 
A multi-agent framework has been proposed in~\cite{GeethaDevasena2012} for model-based test generation in software testing. Agent programs are in charge of exploring a UML model of the code, generating all the scenarios of the if-then-else conditions and branches. 
BDI agents have been tested, with respect to the interaction behaviours of multi-agent systems~\cite{Rehman2015}, or single agents (units) in terms of the correctness of their beliefs (e.g., value combinations), plans (e.g., triggering the correct plan according to the context), and events or messages (e.g., sending them at the right time)~\cite{Zhang2008,Zhang2011,Padgham2013}. 
In this paper we turn the table and introduce BDI agents into the test environment, for intuitive and effective test generation. 

\section{CONCLUSIONS} \label{sc:conclusion}

In this paper, we presented the use of belief-desire-intention (BDI) agents for model-based test generation, in the context of verifying robotic code for HRI in simulation. 
BDI agents allow more realistic human-like stimulus, whilst simultaneously facilitating the generation of interesting events, to gain good coverage levels of the code under test, and the possibilities of the robot and the environment. 

We developed a CDV testbench in the context of HRI, for a robotic simulator in ROS-Gazebo, where a human and a robot collaboratively manufacture a table. 
The use of the CDV methodology simplified the testing processes, through automated test generation, driving of tests, checking assertions to monitor the satisfaction of functional and safety requirements, and coverage collection. 
Tests were derived to stimulate the human and environment, which stimulates the robot's code under test, in an indirect, but more realistic manner. 
The test suite was generated in two tiers, where abstract stimulus sequences are computed first, and then concretized through constrained random sampling. 
We complemented BDI-based test generation with constrained, pseudorandom generated abstract tests. 
Our results highlight the potential of using BDI agents for test generation, stimulating the code according to realistic scenarios, from complex and detailed models of the environment and the robot's components.

In the future, we plan to experiment with the composition and control options of the BDI agents (e.g., using both beliefs and goals). 
Efficient and intelligent forms to control the BDI agents, preferably in an automated manner, need to be explored. 
Potentially, agents can be fused, made to learn, or improve coverage by themselves, among other options. 
Model checking could also be used to explore the BDI multi-agent system. 
Also, we would like to introduce feedback loops to guide the BDI agents to produce effective tests.

\bibliographystyle{plain}
\bibliography{robosafe}

\end{document}